%% file: main.tex
\documentclass[10pt,twocolumn,letterpaper]{article}

\usepackage{iccv}
\usepackage{times}
\usepackage{epsfig}
\usepackage{graphicx}
\usepackage{amsmath}
\usepackage{amssymb}
\usepackage{booktabs}
\usepackage{subfig}
\usepackage{amsmath}
\usepackage{nopageno}
\usepackage[breaklinks=true,bookmarks=false]{hyperref}

\iccvfinalcopy % *** Uncomment this line for the final submission

\def\iccvPaperID{15} % *** Enter the ICCV Paper ID here

\def\resp{\emph{resp.}}
\ificcvfinal\fi
\pdfoutput=1
\begin{document}

%%%%%%%%% TITLE
\title{\ Deep Domain Adaptation by Geodesic Distance Minimization}

\author{Yifei Wang,  Wen Li,  Dengxin Dai,  Luc Van Gool\\
EHT Zurich\\
Ramistrasse 101, 8092 Zurich\\
{\tt\small yifewang@ethz.ch}
{\tt\small \{liwen, dai, vangool\}@vision.ee.ethz.ch}
}

\maketitle
%\thispagestyle{empty}

%%%%%%%%% ABSTRACT
\begin{abstract}
\input{abstract.tex}
\end{abstract}

%%%%%%%%% BODY TEXT
\section{Introduction}
\input{introduction.tex}
\section{Related Work}
\input{related_work.tex}
%-------------------------------------------------------------------------
\section{Methodology}
\input{methology.tex}
%-------------------------------------------------------------------------
\section{Experiment and Discussion}
\input{experiment_result.tex}
%-------------------------------------------------------------------------
\section{Conclusion and Future Work}
\input{conclusion.tex}
%-------------------------------------------------------------------------

\input{main.bbl}
%{\small
%\bibliographystyle{ieee}
%\bibliography{egbib}
%}

\end{document}

%% file: abstract.tex
In this paper, we propose a new approach called Deep LogCORAL for unsupervised visual domain adaptation. Our work builds on the recently proposed Deep CORAL method, which aims to train a convolutional neural network and simultaneously minimize the Euclidean distance of convariance matrices between the source and target domains. By observing that the second order statistical information (\textit{i.e.} the covariance matrix) lies on a Riemannian manifold, we propose to use the Riemannian distance, approximated by Log-Euclidean distance,
%the Log-Euclidean distance, which endows Riemannian manifold, which naturally represents geodesic distance between two covariance matrices
to replace the naive Euclidean distance in Deep CORAL. 
We also consider first-order information, and minimize the distance of mean vectors between two domains. We build an end-to-end model, in which we minimize both the classification loss, and the domain difference based on the first-order and second-order information between two domains. Our experiments on the benchmark Office dataset demonstrates the improvements of our newly proposed Deep LogCORAL approach over the Deep CORAL method, as well as the further improvement when optimizing both orders of information.
% The results show that Deep LogCORAL raise $0.68\%$ of the average accuracy compared to Deep CORAL. Further more we add first order statistic information (\textit{i.e.} the mean) of the two domains along with second order covariance matrices to have a better representation of the geodesic distance and the result is further increased by $0.35\%$.

%% for the submission
%In this paper, we propose a new approach called Deep LogCORAL for unsupervised visual domain adaptation. Our work builds on the recently proposed Deep CORAL method, which proposed to train a convolutional neural network and simultaneously minimize the Euclidean distance of convariance matrices between the source and target domains. We propose to use the Riemannian distance, approximated by Log-Euclidean distance, to replace the naive Euclidean distance in Deep CORAL. We also consider first-order information, and minimize the distance of mean vectors between two domains. We build an end-to-end model, in which we minimize both the classification loss, and the domain difference based on the first and second order information between two domains. Our experiments on the benchmark Office dataset demonstrate the improvements of our newly proposed Deep LogCORAL approach over the Deep CORAL method, as well as further improvement when optimizing both orders of information.

%% file: introduction.tex
One of the most fundamental assumption in traditional machine learning is that the training data and the test data have identical distributions. However, this may not always hold for real-world visual recognition applications. The limitations of collecting training data, and the large variance of test data in real-world applications make it difficult to guarantee that training and test data follow an identical distribution. As a result, the performance of the visual recognition model can significantly drop due to the distribution mismatch between training and test data. This is known as the "domain adaptation" problem~\cite{duan2012visual}~\cite{gong2012overcoming}~\cite{duan2012domain}~\cite{fernando2013unsupervised}~\cite{li2014learning}~\cite{li2017domain}~\cite{tommasi2014testbed}~\cite{saenko2010adapting}~\cite{gong2012geodesic}~\cite{ganin2015unsupervised}~\cite{patel2015visual}. Torrabla and Efros~\cite{torralba2011unbiased} pointed out that each existing visual recognition dataset more or loss has its own bias, which reveals the common existence of visual domain adaptation problems.

In visual domain adaptation, the domain of the training data is referred to as the source domain, and the domain of the test data is referred to as the target domain. Visual domain adaptation aims to reduce the distribution mismatch between these two domains, such that the performance of visual recognition models learned from the source domain can be improved when testing on the target domain. Typically, the source domain contains a large number of labeled data for training the models, whereas the target domain contains only unlabeled data. Visual domain adaptation has attracted more and more attentions from computer vision researchers in recent years. It becomes even more important after the revival of Convolutional Neural Network (CNN), because CNN usually requires a large number of labeled training data to build a robust model, and it can be expensive to annotate a large number of training data which have an identical distribution as the test data.

A few papers have proposed unsupervised visual domain adaptation based on CNNs \cite{ganin2015unsupervised}\cite{tommasi2014testbed}\cite{long2016unsupervised}\cite{ghifary2016deep}\cite{tzeng2017adversarial}. The recent Deep CORAL method was proposed to reduce the domain difference by minimizing the Euclidean distance between the covariance matrices in the source and target domains~\cite{sun2016deep}. They built an end-to-end model, in which they simultaneously minimized the classification loss and the domain difference. While the Deep CORAL method improves the classification performance of CNN, it is still unclear if the naive Euclidean distance a good choice for minimizing the distance of two covraince matrices. Moreover, only the second-order statistical information (\textit{i.e.}, the covariance matrix) is used in Deep CORAL, and other information is discard. 

\begin{figure}[t]
  \includegraphics[width=0.95\linewidth]{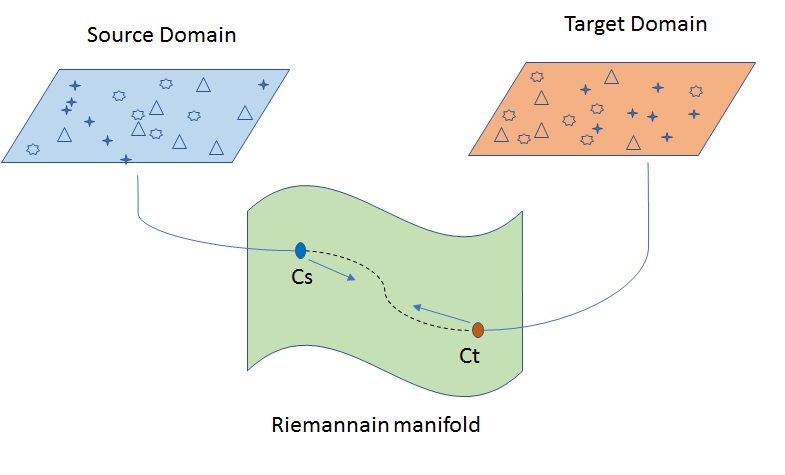}
  \caption{Illustration of our proposed domain adaptation method: minimizing geodesic distance between two domains in Riemannian manifold.}
  \label{fig:log_euclidean}
\end{figure}

% This method shorten the distance between the two domains and therefore the model can work better on test domain. 
% The reasoning behind it seems convincing, however before going deeper into this method we need to ask two crucial questions:
% \textbf{\textit{Is Euclidean distance a good choice for measuring the distance between the two domains' covariance matrices?}}
% \textbf{\textit{Given that we use second order covariance matrices, what is the relationship between first order and second order information and would it be better to use both orders of information to represent the geodesic distance?}}
To cope with the first issue, we propose a new Deep LogCORAL approach which employs the geodesic distance to replace the naive Euclidean distance in Deep CORAL. Intuitively, the covariance matrix is a positive semi-definite (PSD) matrix, which lies on a Riemannian manifold. The source and target covariance matrices can be deemed as two points on the Riemannian manifold, so a more desirable metric is the geodesic distance between the two points on the Riemannian manifold. As inspired by \cite{cui2014flowing}, we employ the LogEuclidean distance, which has been widely used for calculating the geodesic distance between PSD matrices. As shown in Figure~\ref{fig:log_euclidean},
$C_s$ and $C_t$ represent the covariance matrices of the source and target domains, respectively. The crosses, triangles and stars in the blue (\resp, orange) rectangle denote the training samples of different classes from the source (\resp, target) domain. 
By minimizing the Log-Euclidean distance instead of the naive Euclidean distance between these two domains, we expect the domain shift between two domains can be reduced smoothly. We designed a new Log-Euclidean loss, which is integrated into the CNN for end-to-end training.
%Our results showed an improvement compared to optimizing naive Euclidean distance.

To cope with the second issue, we also exploit the first-order statistical information. We propose to minimize the distance between the mean vectors of two domains, which is closely related to the Maximum Mean Discrepancy (MMD) theory. We simultaneously minimize the mean distance and the Log-Euclidean distance, such that both the first-order and the second-order statistical information of two domains become consistent.
% First order information, the mean value, lies on the Euclidean manifold. We did experiment to only optimize the Euclidean distance of the first order mean matrices of the source and target domains. We then try to find the correlation between the mean and LogCORAL losses. Since we project the first order information on Euclidean manifold and second order information on Riemannian manifold, intuitively they represent different perspectives and might not overlap each other. 

We conduct extensive experiments on the benchmark Office dataset, which shows a clear improvement of our newly proposed Deep LogCORAL when compared with the Deep CORAL method. We also demonstrate that both the first-order and the second-order information are necessary for effectively reducing the domain shift.
%Our experiment shows that the mean and LogCORAL losses have very weak correlations and our further experimentation confirms that combining both information leads to better results.

\begin{figure*}[h]
  \centering
  \includegraphics[width=0.90\linewidth]{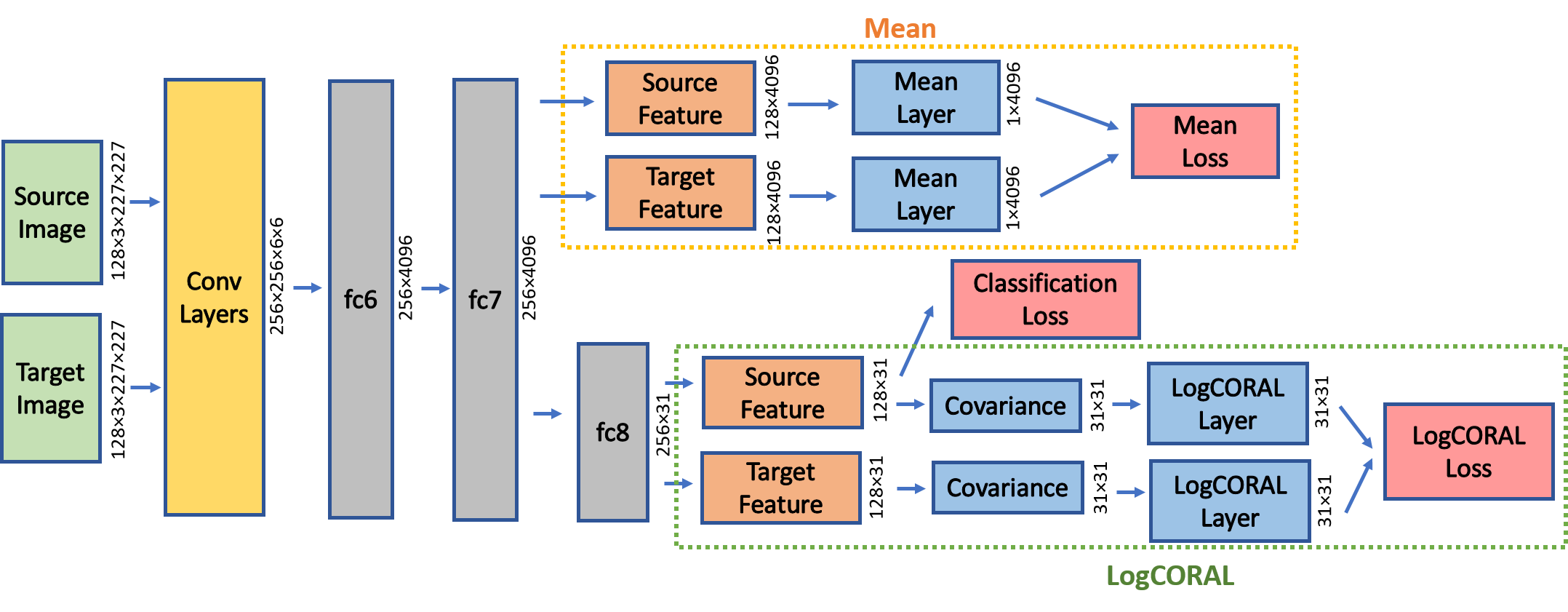}
  \caption{Structure of the model.}
  \label{fig:structure}
\end{figure*}

%% file: related_work.tex
For a comprehension summary for domain adaptation we refer readers to \cite{patel2015visual}.
In \cite{torralba2011unbiased}\cite{tommasi2014testbed} the concept of dataset bias was well introduced and attracted a lot of attention. 
Since then, many methods have been developed in order to overcome the build-in dataset bias. 

Early domain adaptation method like \cite{saenko2010adapting} requires learning a regularized transformation, by using information-theoretic metric learning that maps data in the source domain to the target domain. 
However this method require labeled data from target domain which in many scenarios is unknown to testers. 

Later unsupervised domain adaptation method \cite{gong2012geodesic}\cite{gong2012overcoming}\cite{fernando2013unsupervised}\cite{sun2015return}\cite{cui2014flowing} tried to improve the performance on the target domain by transferring knowledge from the target domain to the source domain without the need of target labels. In \cite{gong2012geodesic} it first extracts features that invariant to domain change then it models the different domains as points on Grassmann manifold and generate number of subspaces in between to train classifier on those subspaces. 
Similarly, to cut back the difference between source domain and test domain \cite{cui2014flowing} generate subspaces in Riemannian manifold and \cite{sun2015return} measure distance in Euclidean distance.
%In \cite{fernando2013unsupervised} source and target domains are represented by subspaces described by eigenvectors and then learning a mapping function which aligns the two domains.

More relevant to this paper is domain adaptation method applied on CNN. The DLID method \cite{chopra2013dlid} is inspired by \cite{gong2012geodesic} to capture information from an "interpolating path" between the source domain and the target domain. Instead of optimizing the representation to minimize some measure of domain shift such as geodesic distance, DRCN\cite{ghifary2016deep} and ADDA \cite{tzeng2017adversarial} alternatively reconstruct the target domain from the source representation.
Gradient reversal method \cite{ganin2015unsupervised} tries to obtain a feed-forward net-work having the same or very similar distributions in the source and the target domains, while RTN \cite{long2016unsupervised} also wants to adapt target classifiers to the source classifiers by learning a residual function.
In \cite{tzeng2014deep} and \cite{long2015learning}, new CNN architectures are proposed, in which \cite{tzeng2014deep} introduces an adaptation layer and an additional main confusion loss to learn a representation while in \cite{long2015learning} all task-specific layers are embedded in a reproducing kernel Hilbert space. 

Our work is mostly inspired by \cite{sun2016deep}, which adds a new CORAL loss that calculates the Euclidean distance of two domain's covariance matrices before the softmax layer. 
This method looks very simple but the result shows it exceeds other methods that appears to be more complex.
Another related paper is about symmetric positive definite (SPD) matrix learning \cite{huang2016riemannian}, that presents a new direction of SPD matrix non-linear learning in the deep neural network model.

%% file: methology.tex
In this section, we present our newly proposed Deep LogCORAL approach. We first give a brief review of the Deep CORAL method, and then introduce our Deep LogCORAL layer as well as the mean layer. 

% Our proposed new approach is build on the Deep CORAL method~\cite{sun2016deep}. For generalization and computational simplicity Deep CORAL used the AlexNet \cite{krizhevsky2012imagenet} pre-trained model. Our work also implement this structure in order to do a fair comparison. 

\subsection{Deep CORAL approach}
As described in \cite{sun2016deep}, CORAL loss is built to calculate the  distance of second-order statistical information between two domains. It first calculates the covariance matrix of the features extracted from the "fc8" layer for each domain, and then minimizes the Euclidean distance of the covariance matrices of two domains.

Formally, let us denote by $\mathbf{D_{S}} = \left[\mathbf{x_{1}},\ldots,\mathbf{x_{n_{S}}}\right]$ as the source domain features that are extracted out of ``fc8" layer, in which $\mathbf{x_{i}}$ is the $i$-th source sample with $d$ being the dimension of features. Similarly, $\mathbf{D_{T}} = \left[\mathbf{u_{1}},\ldots,\mathbf{u_{n_{T}}}\right]$ denotes the target domain features extracted from the ``fc8" layer, in which $\mathbf{u_{i}}$ is the $i$-th target sample. The CORAL loss is then defined as follows,
 \begin{equation} \label{eq1}
 L_{CORAL} = \frac{1}{4d^2}\|\mathbf{C_{S}} - \mathbf{C_{T}}\|^2,
\end{equation}
in which the covariance matrices $\mathbf{C_{S}}$ and $\mathbf{C_{T}}$ are given as follows:
\begin{align}
\mathbf{C_{S}}&=\frac{1}{n_{S} - 1}(\mathbf{D_{S}}^T\mathbf{D_{S}} - \frac{1}{n_{S}}(\mathbf{1}^T\mathbf{D_{S}})^T(\mathbf{1}^T\mathbf{D_{S}}))\\
\mathbf{C_{T}}&=\frac{1}{n_{T} - 1}(\mathbf{D_{T}}^T\mathbf{D_{T}} - \frac{1}{n_{T}}(\mathbf{1}^T\mathbf{D_{T}})^T(\mathbf{1}^T\mathbf{D_{T}}))
\end{align}
where $n_{S}$, $n_{T}$ is the batch size of the source domain and target domain respectively. $d$ is the feature dimension, and $\mathbf{1}^T$ is a vector that all elements equals to $1$.

The Deep CORAL method simultaneously minimizes the above loss and the classification loss, such that the domain distribution mismatch is minimized, and the discriminative ability is also preserved. 

\begin{table*}[h]
\caption{Accuracy comparison for the CNN (without adaptation), CORAL (baseline adaptation) and combine method (extended adaptation method combined with LogCORAL and mean model). Note that A: amazon, W:webcam, D: DSLR, A-W means use amazon as source domain and use webcam as target domain (analogous for the rest).}
\centering
\begin{tabular}{@{}llllllll@{}}
\toprule
Accuracy             & A - W                   & D - W                   & A - D                   & W - D                   & W - A                   & D - A                   & AVG  \\ 
\midrule
CNN                  & 63.34$\pm$0.88          & 95.21$\pm$0.52          & 65.14$\pm$1.26          & 99.26$\pm$0.06          & 49.23$\pm$0.22          & 51.37$\pm$0.48          & 70.59 \\
CORAL                & 66.12$\pm$0.45          & 95.24$\pm$0.49          & 66.38$\pm$2.54          & 99.24$\pm$0.10          & 50.71$\pm$0.24          & \textbf{53.12$\pm$0.69} & 71.80 \\
Ours (LogCORAL+Mean) & \textbf{70.15$\pm$0.57} & \textbf{95.45$\pm$0.07} & \textbf{69.41$\pm$0.51} & \textbf{99.46$\pm$0.31} & \textbf{51.57$\pm$0.46} & 51.15$\pm$0.32          & \textbf{72.87} \\
\bottomrule
\end{tabular}
\label{table:result1}
\end{table*}

\subsection{Deep LogCORAL approach}
To push the classifier closer to target domain, an essential problem is to precisely model the distance between two domains. Deep CORAL use squared Euclidean distance to measure distance, while evidence shows that measure distance on other manifolds such as Riemannian manifold may be more precise in measuring matrix distance and can get better result in domain adaptation \cite{huang2016riemannian}. According to this assumption, we design a new LogCORAL layer after ``fc8" to measure distance in Riemannian manifold.

Covariance matrix is a symmetric positive semi-definite (PSD) matrix, but adding a small $\epsilon$ to the eigenvalues of covariance matrix transforms it into SPD without significantly change its property. Therefore after getting covariance matrices from the source and target domains we can calculate the distance on Riemannian manifold.

\textbf{Log-Euclidean Riemannian metric:} Log-Euclidean metrics was first proposed in \cite{arsigny2007geometric}. It has the capacity to endow Riemannian manifold and also demonstrated that the swelling effect which is clearly visible in the Euclidean case disappears in Riemannian cases. 
Logarithm operation on the eigenvalue of PSD matrices make the manifold to be flat, and then Euclidean distance can be calculated on this flat space, which makes it much easier to caculate the geodesic distance.
 
\textbf{LogCORAL Forward Propagation:} The input of this layer is the covariance matrices calculated in Equation~(2) and (3). The forward process can be easily calculated by using singular value decomposition (SVD) to get the eigenvalues and eigenvectors of the covariance metrics, followed by applying logarithm operation on eigenvalues. The LogCORAL loss is defined as the Euclidean distance between the logarithm of covariance matrices:
\begin{equation} \label{eq4}
% L_{LogCORAL} = \frac{1}{4d^2}\|\mathbf{C_{S}\prime} -\mathbf{C_{T}\prime}\|^2
L_{LogCORAL} = \frac{1}{4d^2}\|log(\mathbf{C_{S}}) -log(\mathbf{C_{T}})\|^2,
\end{equation}
% The modified covariance matrices $\mathbf{C_{S}\prime}$ and $\mathbf{C_{T}\prime}$ are given as follows:
% \begin{align}
% \mathbf{C_{S}\prime}&=log(\mathbf{C_{S}}) = \mathbf{U_{S}}log(\mathbf{\Sigma_{S}})\mathbf{V_{S}}^T\\
% \mathbf{C_{T}\prime}&=log(\mathbf{C_{T}}) = \mathbf{U_{T}}log(\mathbf{\Sigma_{T}})\mathbf{V_{T}}^T
% \end{align}
where the $log()$ operation is the logarithm of the PSD matrix. We take the source domain covariance matrix $\mathbf{C_{S}}$ as an example. Let us denote the eigen-decomposition of $\mathbf{C_{S}}$ as  $\mathbf{C_{S}} = \mathbf{U_{S}}\mathbf{\Sigma_{S}}\mathbf{U_{S}}^T$, then the Logarithm operator is defined as $log(\mathbf{C_{S}}) = \mathbf{U_{S}}log(\mathbf{\Sigma_{S}})\mathbf{U_{S}}^T$, where $log(\mathbf{\Sigma_{S}})$ is calculated by applying the Logarithm operator on the diagonal elements of $\mathbf{\Sigma_{S}}$. The same procedure is applied to the target domain covariance matrix.

\textbf{LogCORAL Back Propagation:} Back-propagation can be derived following the technique described in \cite{huang2016riemannian}. For simplicity, we denote $\mathbf{C_{S}\prime}=log(\mathbf{C_{S}})$ and $\mathbf{C_{T}\prime}=log(\mathbf{C_{T}})$. Taking the source domain as an example, the gradients can be derived as follows:
 \begin{equation} \label{eq7}
 \frac{\partial L_{LogCORAL}}{\partial \mathbf{C_{S}}} = \frac{1}{2d^2}(\mathbf{C_{S}\prime}-\mathbf{C_{T}\prime})\frac{\partial \mathbf{C_{S}\prime}}{\mathbf{C_{S}}},
 \end{equation}
where 
\begin{multline}
\frac{\partial \mathbf{C_{S}\prime}}{\mathbf{C_{S}}}=\mathbf{U_{S}}( \mathbf{P}^T \circ (\mathbf{U_{S}}^Td\mathbf{U_{S}} ))_{sym} \mathbf{U_{S}}^T + \\ \mathbf{U_{S}}(d\mathbf{{\Sigma}_{S}} )_{diag}\mathbf{U_{S}}^T
\end{multline}
\begin{align} \label{eq8}
d\mathbf{U_{S}}&=2( \frac{\partial L_{LogCORAL}}{\partial \mathbf{C_{S}\prime}} )_{sym} \mathbf{U_{S}}log(\mathbf{\Sigma_{S}})\\
d\mathbf{\Sigma_{S}}&=\mathbf{\Sigma_{S}}^{-1} \mathbf{U_{S}}^T(\frac{\partial L_{LogCORAL}}{\partial \mathbf{C_{S}\prime}} )_{sym} \mathbf{U_{S}}\\
\mathbf{P(i,j)}&=\left\{
                \begin{array}{ll}
                  \frac{1}{\sigma_{i}-\sigma_{i}}, i \neq j,\\
                  0,   i=j
                \end{array}
              \right.
\end{align}
where $\circ$ is Hadamard product, \ie, element wise product, $sym$ operation is defined as $\mathbf{A_{sym}} = \frac{1}{2}(\mathbf{A}+\mathbf{A}^T)$, $diag$ operator is to keep only  diagonal values of $\mathbf{A}$  and set the rests to zeros, and  $\sigma_{i}$ denotes the $i$-th eigenvalue in $\mathbf{\Sigma_{S}}$.
For the target domain, the calculation is the same except the sign is negative.

After implementing the LogCORAL layer, we can build our Deep LogCORAL structure. As in Deep CORAL, we have source domain with label and target domain without label as shown in the green rectangular in Figure~\ref{fig:structure}. 

\subsection{Mean Layer}
Note that we have considered the second-order statistic information between two domains, intuitively we can also use the first order statistic information, the mean value, as it is also one type of the representative information for a dataset. As shown in the green rectangular of Figure~\ref{fig:structure}. We create a mean layer after fc7 which calculates the mean value along the column to get a mean vector of features. Then we calculate the Euclidean distance of the mean vectors between two domains as mean loss.
Definition is showen as follows:
\begin{equation} \label{eq12}
\mathbf{L_{Meanloss}} = \frac{1}{2d}\|\mathbf{1}^T\mathbf{D_{S}} - \mathbf{1}^T\mathbf{D_{T}}\|^2
\end{equation}
 
By incorporating the mean loss into the model, now there are three losses: classification, LogCORAL and mean losses to be optimized. The whole structure is shown in Figure~\ref{fig:structure}.

\begin{table*}[h]
\caption{Accuracy comparison for the Mean, LogCORAL and combined model (combined with LogCORAL and mean model).}
\centering
\begin{tabular}{@{}llllllll@{}}
\toprule
Accuracy      & A - W                   & D - W                   & A - D                   & W - D                   & W - A                   & D - A                   & AVG  \\ 
\midrule
Mean          & 66.29$\pm$0.74          & \textbf{95.56$\pm$0.19} & 68.67$\pm$0.46          & 99.51$\pm$0.23          & 49.83$\pm$0.85          & 50.74$\pm$0.74          & 71.77 \\
%CORAL         & 66.12$\pm$0.45          & 95.24$\pm$0.49          & 66.38$\pm$2.54          & 99.24$\pm$0.10          & 50.71$\pm$0.24          & \textbf{53.12$\pm$0.69} & 71.80 \\
%CORAL+Mean    & 67.51$\pm$0.83          & 95.43$\pm$0.43          & 67.86$\pm$0.95          & 99.36$\pm$0.34          & 50.19$\pm$0.57          & 51.13$\pm$0.96          & 71.94 \\
LogCORAL      & 68.83$\pm$0.57          & 95.23$\pm$0.15          & 68.64$\pm$1.41          & \textbf{99.52$\pm$0.41} & 50.94$\pm$0.28          & \textbf{51.73$\pm$0.61} & 72.48 \\
LogCORAL+Mean & \textbf{70.15$\pm$0.57} & 95.45$\pm$0.07          & \textbf{69.41$\pm$0.51} & 99.46$\pm$0.31          & \textbf{51.57$\pm$0.46} & 51.15$\pm$0.32          &\textbf{72.87}\\
\bottomrule
\end{tabular}
\label{table:result2}
\end{table*}

%\begin{table*}[h]
%\caption{Accuracy comparison for our model and other state of the art methods.}
%\centering
%\begin{tabular}{@{}llllllll@{}}
%\toprule
%Accuracy                        & A - W                   & D - W                   & A - D                   & W - %D                   & W - A                   & D - A                   & AVG  \\ 
%\midrule
%DAN\cite{long2015learning}      & 68.5$\pm$0.4            & 96.0$\pm$0.3            & 67.0$\pm$0.4            & %99.0$\pm$0.2            & 53.1$\pm$0.3            & 54.0$\pm$0.4            & 72.9 \\
%GRL\cite{ganin2015unsupervised} & 72.6$\pm$0.3            & 96.4$\pm$0.1            & 67.1$\pm$0.3            & %99.2$\pm$0.3            & 52.7$\pm$0.2            & 54.5$\pm$0.4            & 73.7 \\
%JAN\cite{long2016deep}          & 74.9$\pm$0.3            & \textbf{96.6$\pm$0.2}   & 71.8$\pm$0.2            & %99.5$\pm$0.2            & 55.0$\pm$0.4            & \textbf{58.3$\pm$0.3}   & 76.0\\
%JAN+A\cite{long2016deep}        & \textbf{75.2$\pm$0.4}   & \textbf{96.6$\pm$0.2}   & \textbf{72.8$\pm$0.3}   & %\textbf{99.6$\pm$0.1}   & \textbf{56.3$\pm$0.2}   & 57.5$\pm$0.2            & \textbf{76.3}\\
%Ours (LogCORAL+Mean)            & 70.2$\pm$0.6            & 95.5$\pm$0.1            & 69.4$\pm$0.5            & %99.5$\pm$0.3            & 51.6$\pm$0.5            & 51.2$\pm$0.3            & 72.9 \\
%\bottomrule
%\end{tabular}
%\label{table:result3}
%\end{table*}

%% file: experiment_result.tex
%\begin{figure*}[h]
%   \centering
%   \subfloat[]{\includegraphics[width=0.2\textwidth]{images/cnn_amazon1.png}}
%   \hfill
%   \subfloat[]{\includegraphics[width=0.2\textwidth]{images/cnn_webcam1.png}}
%   \hfill
%   \subfloat[]{\includegraphics[width=0.2\textwidth]{images/log_amazon1.png}}
%   \hfill
%   \subfloat[]{\includegraphics[width=0.2\textwidth]{images/log_webcam1.png}}
%   \caption{Comparison of the covariance matrix of CNN and Deep LogCORAL on domain shift A-W. (a):Source domain covariance matrix using CNN features. (b):Target domain covariance matrix using CNN features. (c):Source domain covariance matrix using Deep LogCORAL features. (d):Target domain covariance matrix using Deep LogCORAL features.} 
%   \label{fig_covanrice}
%\end{figure*}
%\begin{figure*}[h]
%   \centering
%   \subfloat[]{\includegraphics[width=0.2\textwidth]{images/cnn_amazon_mean.png}}
%   \hfill
%   \subfloat[]{\includegraphics[width=0.2\textwidth]{images/cnn_webcam_mean.png}}
%   \hfill
%   \subfloat[]{\includegraphics[width=0.2\textwidth]{images/mean_amazon_mean.png}}
%   \hfill
%   \subfloat[]{\includegraphics[width=0.2\textwidth]{images/mean_webcam_mean.png}}
%   \caption{Comparison of the mean matrix of CNN and mean model on domain shift A-W. (a):Source domain mean matrix using CNN features. (b):Target domain mean matrix using CNN features. (c):Source domain mean matrix using mean model features. (d):Target domain mean matrix using mean model features.} 
%   \label{fig_mean}
%\end{figure*}

\subsection{Experimental settings}
For a fair comparison, we follow the experimental setting in Deep CORAL, \textit{i.e.} using the Office dataset and the ImageNet pretrained AlexNet model. The Office dataset contains images collected from three domains: \textit{Amazon, DSLR, Webcam}, each has identical thirty one categories.  We use one of the three domains as the source domain, and one of the rest two as the target domain, leading to six cases in total.

To make the training procedure more stable, moving average is employed when computing the losses. We use accumulated covariance and mean value to calculate LogCORAL loss and mean loss:
\begin{equation}
\mathbf{C}=0.9 * \mathbf{\widetilde{C}} + 0.1 * \mathbf{C_{batch}}
\end{equation}
\begin{equation}
\mathbf{M}=0.9 * \mathbf{\widetilde{M}} + 0.1 * \mathbf{M_{batch}},
\end{equation}
where $\mathbf{C}$ is the moving average covariance matrix, $\mathbf{\widetilde{C}}$ is the accumulated covariance from last iteration, and $\mathbf{C_{batch}}$ is the covariance matrix from the current batch. The terms for mean value are similarly defined as above.

\subsection{Experimental results}
We compare our proposed model with the Deep CORAL method, and also include the CNN method as a baseline, in which we fine-tune the pretrained AlexNet model using source domain samples without considering domain adaptation. 
% For each domain shift we test six different models: CNN (only optimize classification loss), Deep CORAL (optimize classification loss and CORAL loss), Mean model (optimize classification loss and mean loss), Deep LogCORAL (optimize classification loss and LogCORAL loss), combination model of CORAL and Mean (optimize classification loss, mean loss and CORAL loss) and combination model of LogCORAL and Mean (optimize classification loss, mean loss and LogCORAL loss). 

The experimental results are shown in Table~\ref{table:result1}. After applying our combination model (LogCORAL+Mean), for each domain shift, five out of six shifts reach the highest accuracy. The average accuracy raised $2.28\%$ and $1.07\%$ compared to CNN and Deep CORAL. 

We further conduct an ablation study as shown in Table~\ref{table:result2}. Optimizing the mean loss or the LogCORAL loss individually could gain performance improvements over the baseline CNN method, which shows the first-/second-order information is useful for domain adaptation. We also observe that optimizing LogCORAL outperforms the Deep CORAL method in Table 1 in terms of average accuracy, which demonstrates the effectiveness of minimizing geodesic distance instead of using simple Euclidean distance on covariance matrices. Combing the mean loss and logCORAL loss gives further improvements in general, and raises the average accuracy by $0.39\%$. 

\begin{figure}[h]
  \centering
  \includegraphics[width=1.0\linewidth]{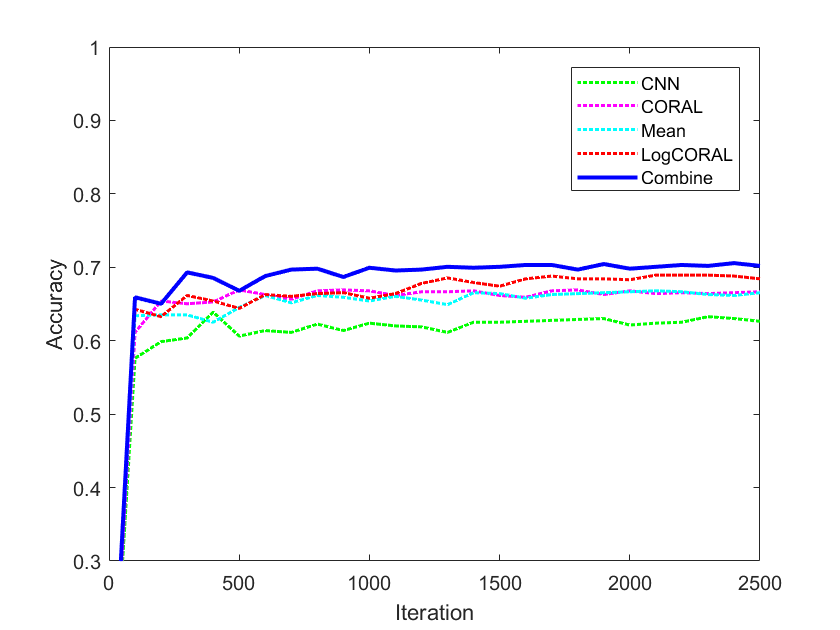}
  \caption{Comparison of learning curve of the models on A-W domain shift.}
  \label{fig:comp}
\end{figure}

We take A-W as an example to show the learning curves in Figure~\ref{fig:comp}. We observe that our proposed model converge fast. Moreover, minimizing the mean loss or the Deep LogCORAL loss individually improves the test accuracy compared to the baseline CNN method, and combining two losses further improves the accuracy.
Furthermore, we also conduct additional experiment by combining the mean loss with the Deep CORAL loss, which improves the average accuracy from  $71.80\%$ to $71.94\%$. This again verifies our motivation that it is beneficial to use both first-order and second-order statistical information for domain adaptation. However, this result is still worse than our final approach (logCORAL+Mean). We attribute this to the usage of geodesic distance on second order covariance matrices in Riemannian space which has only weak correlation with first order Euclidean information. We will demonstrate this in the following session.

% Optimizing mean loss has almost the same performance as Deep CORAL, while the combined model achieves better result in general. LogCORAL+Mean raised $1.10\%$ accuracy compared to optimize mean loss alone. However when applying the same to integrate CORAL and mean losses the improvement is very limited, only raised $0.17\%$. 
% As mean loss and CORAL loss are both Euclidean distances , optimizing these two losses at the same time makes the model redundant because of the information overlapping. The reason that LogCORAL+Mean has a more obvious improvement will be explained later by showing the weak correlation between Euclidean distance and Riemannian distance.

\subsection{Visualization}
%To visualize how Deep LogCORAL model minimizes the distance between two domains, we plot the covariance matrices of source (amazon) and target (webcam) domains after feeding two domain images in CNN and Deep LogCORAL respectively, see Figure~\ref{fig_covanrice}. Clearly distance is shorter after optimize LogCORAL loss. 
%Similarly, to visualize how mean loss works, we plot the mean matrices of the two domains after training CNN and Mean model. Because the the mean matrix dimension is $1\times4096$ so we reshape it in a $64\times64$ matrix to have better visualization. Also the distance shows to be shorter: the distribution of the data in source and target domain is more similar after optimizing the mean loss. See Figure~\ref{fig_mean}.
After demonstrating that minimizing the LogCORAL and mean losses individually helps shorten the distance between the two domains, we further investigate if the Euclidean and Riemannian distances correlated to each other. 

 %Intuitively, both try to minimize the difference between two representative distance matrices under certain distance metric, they might be irrelevant.
\begin{figure*}[h]\label{fig2}
\centering
  \subfloat[]{\includegraphics[width=0.46\linewidth]{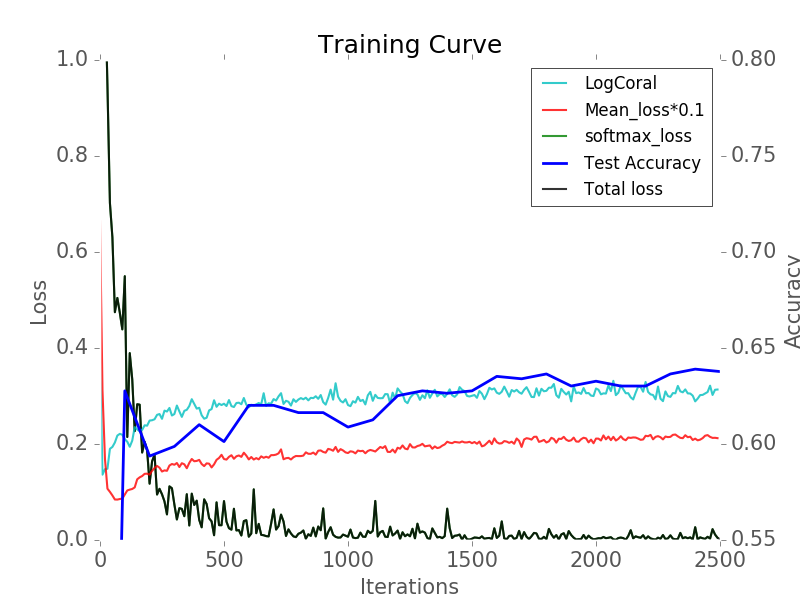}\label{fig2:f1}}
  \centering
  \subfloat[]{\includegraphics[width=0.46\linewidth]{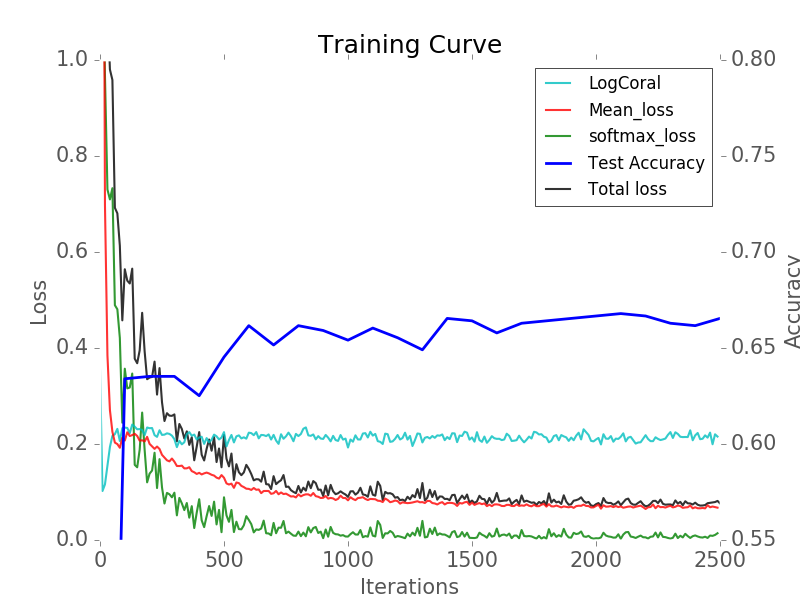}\label{fig2:f2}}
   \hfill
  \centering
  \subfloat[]{\includegraphics[width=0.46\linewidth]{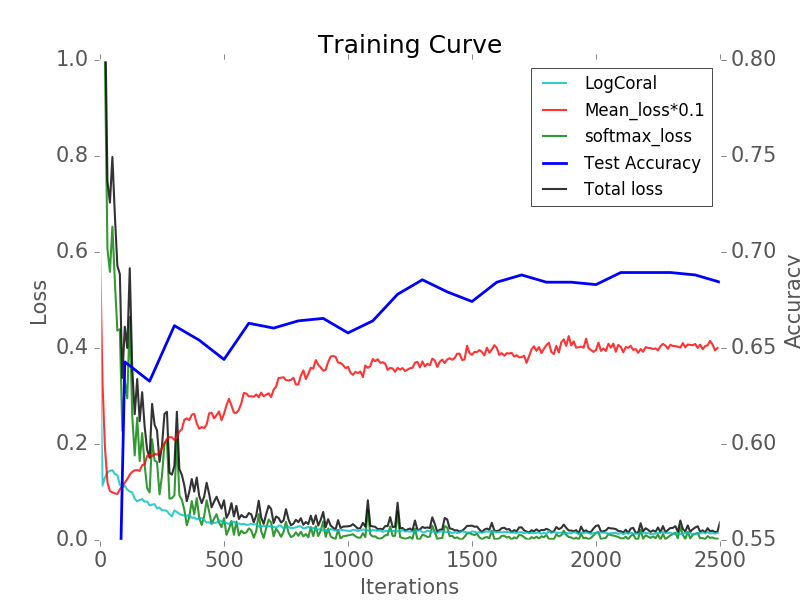}\label{fig2:f3}}
  \centering
  \subfloat[] {\includegraphics[width=0.46\linewidth]{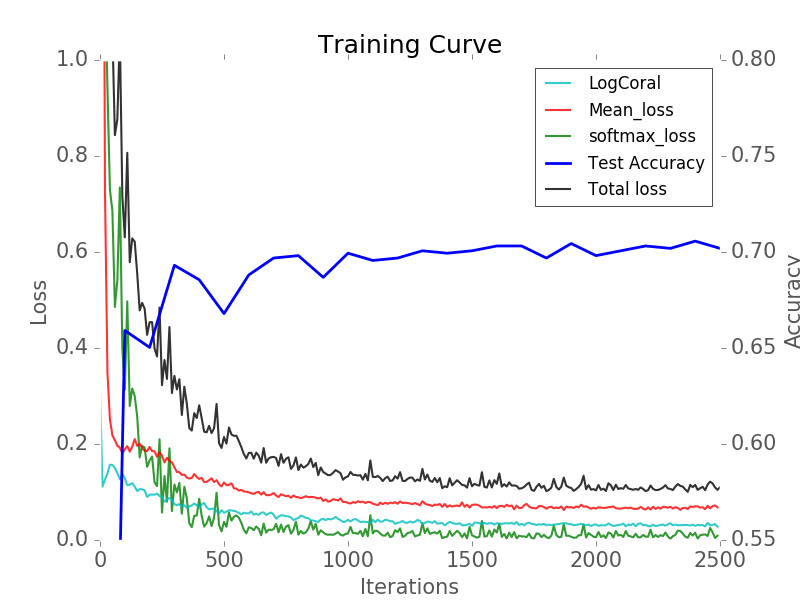}\label{fig2:f4}}
  \caption{learning curves of different models on domain shift A-W. (a):Learning curve of CNN (\textit{i.e.} only optimize classification loss). (b):Learning curve of mean model (\textit{i.e.} optimize mean loss and classification loss). (c):Learning curve of Deep LogCORAL (\textit{i.e.} optimize LogCORAL loss and classification loss). (d):Learning curve of combination model (\textit{i.e.} optimize mean loss, LogCORAL loss and classification loss).} 
\end{figure*}

Figure~\ref{fig2:f1} shows the learning curve of CNN, and also calculate the mean and LogCORAL loss even they are not used in learning procedure. Figure~\ref{fig2:f2} shows the learning curve of mean model (\textit{i.e.} optimizing mean loss and classification loss), and we also calculate the LogCORAL loss. 
Figure~\ref{fig2:f3} shows the other way around which optimizes LogCORAL loss and classification loss, and we also calculate the mean loss. 

From Figure~\ref{fig2:f1} we observe that, without optimizing any of those distances, mean loss and LogCORAL loss would both go up. When optimizing mean loss, LogCORAL loss remain stable while mean loss has a obvious drop down, see Figure~\ref{fig2:f2}. However if we optimize LogCORAL loss in Figure~\ref{fig2:f3}, LogCORAL loss goes down but mean loss goes up. Those results indicate that those two losses have very weak correlation. This also explains why minimizing the two distances at the same time can achieve even better accuracy for domain adaptation.

We further show the learning curve in Figure~\ref{fig2:f4}, where we optimize both losses. In this case, both the mean and LogCORAL losses go down, and we achieve better result.

\subsection{Discussion with other state-of-the-arts}
Deep domain adaptation is a fast growing research area. Many state of the art methods have been proposed on this topic. Generally speaking those methods can be classified into two main categories. First category is discrepancy based model, for example DAN \cite{long2015learning}, which directly minimizes the domain discrepancy (\textit{eg}, MMD) to bring the source and target domains closer. The second category is the adversarial model, for example GRL \cite{ganin2015unsupervised}, which aims to confuse a domain classifier to learn transferable features. 

Our method falls into the first category. We extend the CORAL method by using second order Log-Euclidean distance and combining with first order mean loss. We show that a proper distance is important for employing second-order statistical information to minimize the domain discrepancy. 
Some recent works give even higher accuracy on the Office dataset by using both the discrepancy and adversarial principles (for example, the JAN+A method~\cite{long2016deep}), and we believe it would be interesting to incorporate the proposed Log-Euclidean distance into those works to further boost the performance. We leave this to the future work for further study. 

% Our method show certain disadvantage in accuracy improvement compare to other state of the art methods. For the future improvement, it might be possible to further improve the accuracy if we also apply LogCORAL layer after other layers such as fc6 and fc7 or applying the similar technique to minimize joint LogCORAL loss as shown in \cite{long2016deep}. Besides integrating our model with the other state of the arts methods may also be a interesting topic to work on in the future.
% Up to now the highest accuracy achieved on Office detest under the same experiment setting as in this paper is JAN+A \cite{long2016deep} for an accuracy of $76.3\%$ on average. 

%% file: conclusion.tex
In this paper, we proposed a new Deep LogCORAL method to minimize the geodesic distance on Riemannian manifold. We used the Log-Euclidean distance to replace the Euclidean distance in the Deep CORAL method, and also proposed a mean distance to additionally exploit the first-order satistical information for domaina adaptation.  Our experimental results showed that our new Deep LogCORAL method generally outperformed the deep LogCORAL method for unsupervised domain adaptation using the benchmark Office dataset. In the future, we would like to incorporate the proposed LogCORAL loss into more models to futher improves the existing state-of-the-art methods. 